\documentclass[10pt,twocolumn,letterpaper]{article}

\usepackage{cvpr}
\usepackage{times}
\usepackage{epsfig}
\usepackage{graphicx}
\usepackage{amsmath}
\usepackage{amssymb}
\usepackage{amsfonts}
\usepackage{float}
\usepackage{algorithm}
\usepackage[noend]{algpseudocode}
\usepackage{subfigure}
\usepackage{wrapfig}
\usepackage{multirow}

\usepackage{amsthm}
\newtheorem{prob}{Problem}
\newtheorem{rem}{Remark}
\usepackage[breaklinks=true,bookmarks=false]{hyperref}
\usepackage{color}
\usepackage{colortbl}
\definecolor{Gray}{gray}{0.9}
\newcolumntype{a}{>{\columncolor{Gray}}c}
\cvprfinalcopy 

\def\cvprPaperID{3201} 

\ifcvprfinal\fi
\begin{document}
\newcommand{\tcb}[1]{\textcolor{blue}{#1}}
\newcommand{\tcr}[1]{\textcolor{red}{#1}}

\title{A Reinforcement Learning Approach to the View Planning Problem}

\author{Mustafa Devrim Kaba\thanks{These authors contributed to this paper equally.} \and Mustafa Gokhan Uzunbas\footnotemark[1] \and Ser Nam Lim\\
General Electric Global Research Center, 1 Research Circle, Niskayuna, NY 12309.\\
{\tt\small \{mustafa.kaba, mustafa.uzunbas, limser\}@ge.com}
}

\maketitle

\begin{abstract}

We present a Reinforcement Learning (RL) solution to the view planning
problem (VPP), which generates a sequence of view points that are
capable of sensing all accessible area of a given object represented
as a 3D model. In doing so, the goal is to minimize the number of view
points, making the VPP a class of set covering optimization problem (SCOP).
The SCOP is $NP$-hard, and the inapproximability results tell us that
the greedy algorithm provides the best approximation that runs in
polynomial time. In order to find a solution that is better than the
greedy algorithm, (i) we introduce a novel score function by
exploiting the geometry of the $3D$ model, (ii) we model an intuitive
human approach to VPP using this score function, and (iii) we cast VPP
as a Markovian Decision Process (MDP), and solve the MDP in RL
framework using well-known RL algorithms. In particular, we use SARSA,
Watkins-Q and TD with function approximation to solve the MDP. We
compare the results of our method with the baseline greedy algorithm
in an extensive set of test objects, and show that we can out-perform
the baseline in almost all cases.

\end{abstract}

\section{Introduction}
\label{sec:intro}

\begin{figure}[t!]
\begin{center}
\subfigure[]
{
\includegraphics[width=0.3\linewidth]{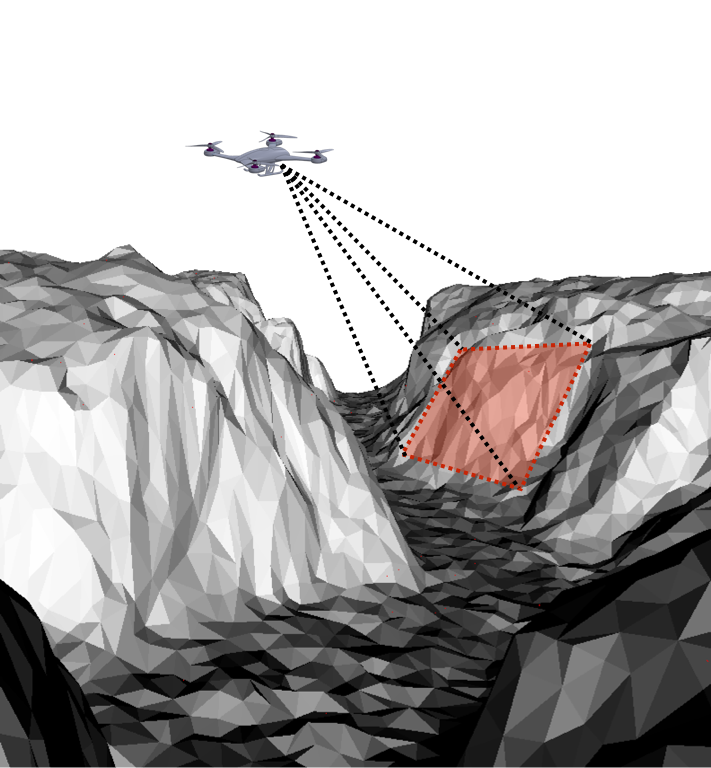}
}
\subfigure[]
{
\includegraphics[width=0.3\linewidth]{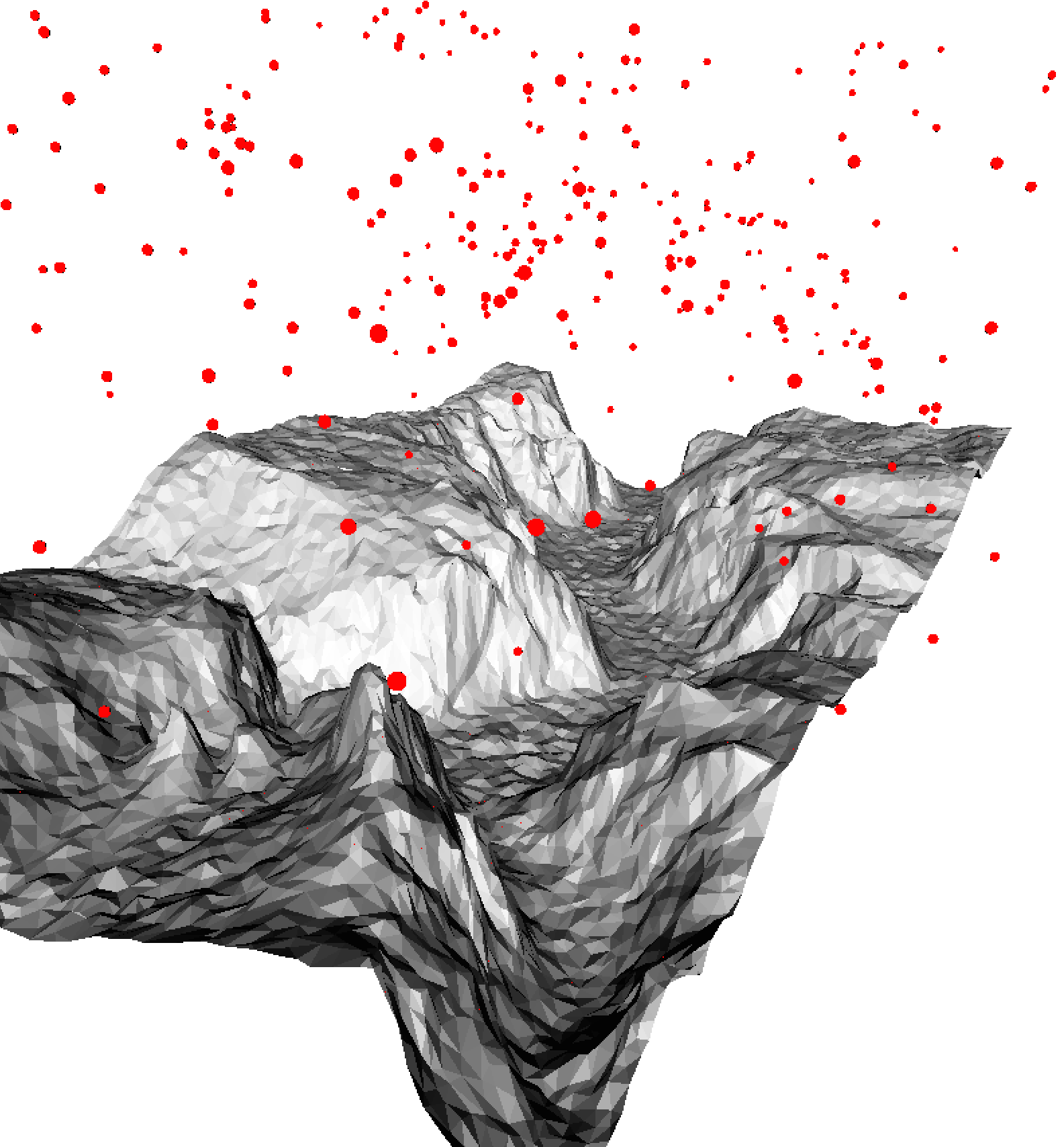}
}
\subfigure[]
{
\includegraphics[width=0.3\linewidth]{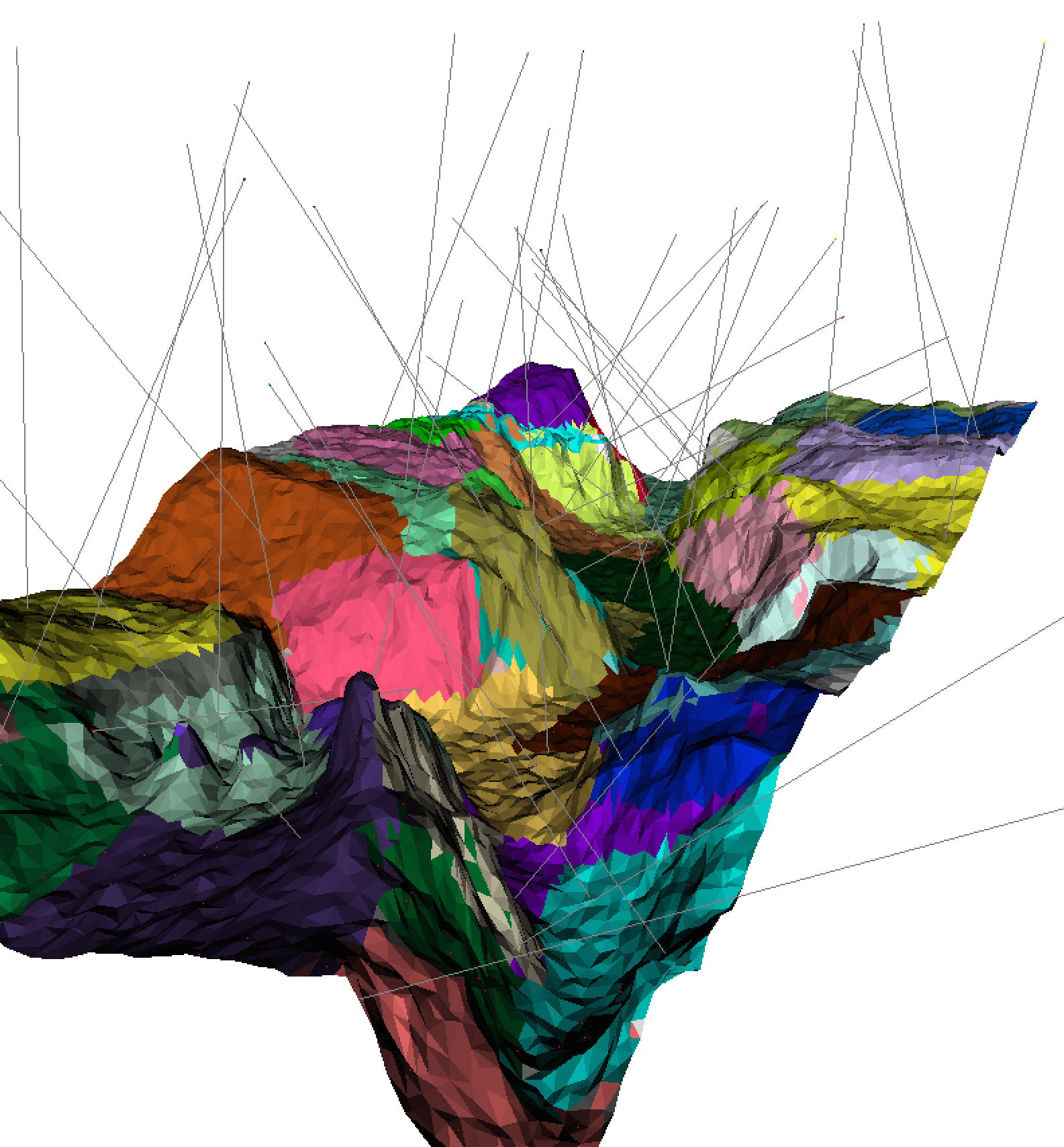}
}
\end{center}
\caption{(a) View planning for UAV terrain modeling, (b) Given a set of initial view points, (c) The goal is to find minimum number of views that provide sufficient coverage. Here, color code represents correspondence between selected views and the coverage.}
\label{fig:goal}
\end{figure}

In this work, we present a solution to the view planning problem
(VPP), which aims to automatically determine a minimum number of
camera perspectives for viewing a given object in order to achieve a
coverage requirement. View planning is becoming increasingly important as the
advent of autonomous platforms is placing demand on developing
algorithms that can provide such a solution, particularly for robots
and UAVs mounted with cameras whose missions are to collect imageries
that fully cover the object of interest (See Figure~\ref{fig:goal}).
In this paper, we will focus on model-based view planning where an
object's 3D model is available. In model-based view planning, one can
take a more global view of the optimization problem involved. This can
be seen in Sheinin~\etal's recent work~\cite{Sheinin_2016_CVPR} that
tries to take a rough $3D$ underwater sonar model and exploit an
optimization criterion that is based on information gain, optimizing
viewpoints so that the descattered albedo is least noisy. In contrast,
non-model-based view planning \cite{4270361, 5597268} often relies on
stochastic state analysis, utilizing uncertainty estimation to plan
the next best view (NBV).

One of the earliest applications of view planning was indoor and
outdoor surveillance, which is also known as the \emph{art gallery
problem} \cite{Mi:08}. More recently, there has been an increased
interest in the use of drones in surveillance, inspection and $3D$
reconstruction, all of which require view planning \cite{MoRu:16,
Ma:99, Wh:97, Mo:14, RaMe:15, Sa:15, SchKo:12, ScRo:03}. In many of
these applications, prior 3D models are available. For example, in
rescue missions, it is critical that survivors be found quickly, and
often such search and rescue missions are conducted from the air. 3D
models of search regions are often readily available (such as from
Google Earth), and can be exploited in view planning to plan the
search paths.

\begin{figure*}[!ht]
\centering
\setlength\tabcolsep{-0.5cm}
\begin{tabular}{ccccccccccc}
cam 1 & cam 2 & cam 3 & cam 4 & cam 5& cam 6& cam 7& cam 8& cam 9&cam 11 &cam 13 \\ \hline
\includegraphics[height=.15\linewidth]{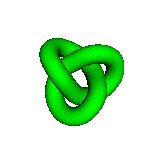} & 
\includegraphics[height=.15\linewidth]{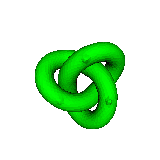} &
\includegraphics[height=.15\linewidth]{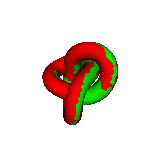} &
\includegraphics[height=.15\linewidth]{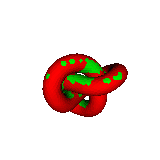} &
\includegraphics[height=.15\linewidth]{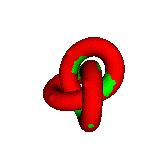} &
\includegraphics[height=.15\linewidth]{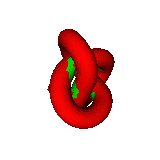} &
\includegraphics[height=.15\linewidth]{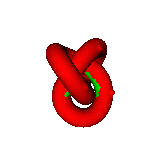} &
\includegraphics[height=.15\linewidth]{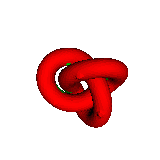} &
\includegraphics[height=.15\linewidth]{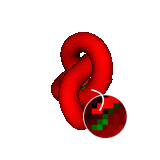} &
    \includegraphics[height=.15\linewidth]{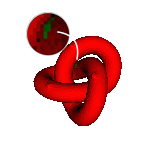}&
    \includegraphics[height=.15\linewidth]{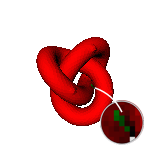} 
    \end{tabular}
    \vspace{-0.1in}
    \caption{Illustration of the bad performance of greedy algorithm in the VPP. Green color represents regions covered by a single camera only, while red color represents regions covered by multiple cameras at different times. The greedy algorithm returns a solution with $13$ cameras. However, the contribution of the last a few cameras are, in fact, minor.}
\label{fig:greedyFails}
\end{figure*}

Model based VPP can be regarded as a set covering optimization problem
(SCOP) and is constrained by the limitations of SCOP \cite{Ta:95}.
Under reasonable complexity assumptions, the na\"{i}ve greedy
algorithm is essentially the best polynomial time approximation
algorithm to the $NP$-hard SCOP \cite{Fe:98}. Even though one can
often find a better solution specific to the problem in hand, to the
best of our knowledge, there is no generic method which is guaranteed
to out-perform the solution provided by the greedy algorithm.
Therefore, we will use the greedy algorithm as a benchmark for the
VPP. On the other hand, greedy algorithm can easily fail and cannot
guarantee the optimal solution to a generic coverage problem. Figure
\ref{fig:greedyFails} illustrates an example of bad performance of the
greedy algorithm. In this example, a $3D$ knot model is covered with a
virtual camera from multiple view points. In the figure, color code
represents areas covered by single (green) and multiple (red) cameras.
Even though the first few cameras effectively increase the coverage,
the last few of them are needed only to cover very small areas that
remained uncovered (magnified in the figure). In this work we propose
an intelligent planning scheme which is capable of reducing this
redundancy.

\begin{wrapfigure}{l}{0.45\linewidth}
\centering
\includegraphics[width=\linewidth, trim = 10 0 10 0, clip]{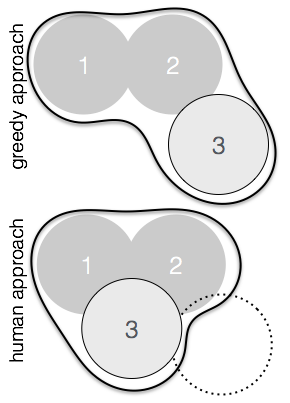} 
\vspace{-0cm}
\caption{Illustration of the purely greedy vs. the intuitive human approach to the set coverage optimization problem. Non-greedy intermediate steps chosen by human leads to a more efficient solution.}
\label{fig:g_vs_h}
\vspace{0pt}
\end{wrapfigure}
In particular, we show that even though the VPP is a set covering
optimization problem, the geometric structure of $3D$ models opens a
path to a more flexible treatment of VPP.
To this end we propose a new set cover score function which allows us to switch between the greedy and non-greedy steps. 
The score function achieves this by penalizing long circumferences if needed. We show that this new scoring scheme can be used to
model the human approach to VPP. We claim that if a human was asked to
solve this problem, s/he would avoid proceeding greedily at certain
steps along the way (See Figure~\ref{fig:g_vs_h}), and this would
eliminate the use of excess view points. 

Choosing between greedy and non-greedy actions at each step
intelligently requires a sequential decision making process which
takes the future actions into account. This essentially converts VPP
to a Markov Decision Process (MDP). The standard way of solving such
MDPs are dynamic programming and reinforcement learning (RL).
Therefore, we employ a RL framework where an agent learns which
actions to take by considering its future consequences. More
specifically, our RL agent learns how to set the parameter of our new
score function at each stage of the coverage task. We implement three
RL algorithms which are mainly built around learning a value function.
More precisely, we use SARSA and Watkins-Q algorithms, which learn the
action value function, and TD algorithm which learns the state value
function \cite{SuBa:98}.

A typical VPP has a large number of initial view points, which induces a MDP with a very large number of states, which in return, necessitates the use of function approximation in RL framework. Hence, we couple the above mentioned algorithms with a nonlinear function approximation scheme.

\subsection*{Our contributions:} 
\begin{itemize}
    \item By exploiting the geometry, we propose a novel, fully automated RL method to solve VPP.
    \item We define a new set coverage score function that can be used to model the human approach to VPP.
    \item With sufficient exploration and learning time, our RL based method provides a solution which is guaranteed to perform at least as good as the greedy algorithm. 
\end{itemize}  

\section{Related Work}

Existing methods that propose solutions to VPP are mainly divided into
two groups: model-based and non-model-based. Non-model-based view
planning differs from the former as the target environment is not
fully observable and is out of the scope of this paper. In this work
we constrain ourselves to model-based view planning, and assume that a
$3D$ CAD model of the environment is already available. In the
literature the model-based view planning is divided into two parts.
The first part is the process of finding the \emph{best} view
locations to cover the object, and the second part is the planning of
the optimal path which includes visiting these selected locations.
The second part is essentially a \emph{Traveling Salesman Problem}
(TSP), and it also remains out of the scope of this work. We
exclusively refer to the first part when we mention VPP. However, it
is worth noting here that an efficient solution to the first part is
very crucial, as it effectively decreases the size of TSP which has to
be tackled in the second part. 

Detailed surveys of proposed solutions to VPP can be found in
\cite{TaATs:95, NeJa:95, ScRoRi:03}. In particular, \cite{TaATs:95}
summarizes the efforts in VPP for inspection, recognition and
reconstruction, \cite{NeJa:95} covers the work on VPP for inspection,
and finally \cite{ScRoRi:03} addresses the VPP problem as it appears
in reconstruction and inspection problems. Among all notable work
studying VPP, our work is in the sprit of the seminal work of Tarbox
and Gottschlich \cite{Ta:95}. Tarbox and Gottschlich also identify the
phenomena portrayed in Figure~\ref{fig:greedyFails} as the cause of
the non-optimality of the greedy algorithm. However, our proposed
solution to handle this problem differs substantially from what they
suggested, namely randomized search with simulated annealing. In the
VPP literature, greedy algorithm is still the most commonly used
algorithm for view point selection \cite{Sc:09, BlAl:07}. There is a
couple of recent work that uses more sophisticated methods such as
linear programming relaxation and genetic algorithms for full $3D$
model coverage \cite{englot2011planning, martin2015evolutionary}.
However, they don't necessarily suggest performance gains over the
greedy algorithm. Lastly, to the best of our knowledge there is no
method in the literature which uses a RL based approach to solve VPP.
However, recently there has been a rapidly growing interest in
combining RL techniques with computer vision. Although they are not
related to VPP, for completeness, we would like to mention
\cite{MnHe:14, Haque_2016_CVPR, Mathe_2016_CVPR, MnKa:13} as the most
recent notable works, which mainly combine deep networks and RL for
digit classification, object detection, person identification and
playing arcade games, respectively.

\section{Problem Formulation}
\label{sec:problem_formulation}

In this paper, we study the view planning problem for $3D$ models.
Without loss of generality, the $3D$ models we consider are triangular
meshes, although other type of meshes could be used as well. We
process models of various objects like geographical terrains, big
structures, interesting geometrical
objects or even machine parts. We formally define the \emph{view
planning problem for $3D$ meshes} as follows
\begin{prob}
Given a $3D$ mesh model $\Omega$ of an object and a finite set of view points ($\ell_i$) together with associated directions ($d_i$), 
$\mathbb{S} := \{(\ell_i, d_i)\}$,
find a subset $\mathbb{T} \subset \mathbb{S}$ of minimum size such
that if identical cameras are placed in locations and in the
directions provided by $\mathbb{T}$, then $\Omega$ can sufficiently be
covered by these cameras. 
\label{prob:definition}
\end{prob}
We unify the two cases where multiple cameras or a single moving
camera is employed and we treat them simultaneously.

\section{Notation and Background}
First, we summarize the mathematical notation that is used. For a given set $Y$, we will denote the power set of $Y$, i.e. the collection of all subsets of $Y$, by $2^Y$. The set of non-negative real numbers will be denoted by $\mathbb{R}_{\geq 0}$. We denote the triangular mesh of interest with $\Omega$. Then, each element of $2^\Omega$ will be a submesh and we denote a submesh (possibly arising from coverage of a single or non-singleton set of views) by $X$.

\subsection{Set covering optimization problem}
Given a set $S$ with finite number of elements, and a collection $\{S_i\}_{i\in I} \subseteq 2^S$ of subsets of $S$ indexed by $I$,
the \emph{set covering optimization problem} is the problem of finding a subset $J$ of $I$ with smallest number of elements satisfying
$
S = \bigcup_{j\in J}S_j.
$
This problem is known to be $NP$-hard, and approximate solutions such as greedy that run in polynomial time are well known, \cite{lin2005near}. 
However, in many instances, it has also been shown that greedy algorithm can not provide the optimal solution, \cite{Erdem:2006db}. Nevertheless, under reasonable assumptions, the inapproximability results of \cite{Fe:98} and \cite{dinur2014analytical} show that the greedy algorithm is the best polynomial-time approximation algorithm one can hope for.

The view planning we posed in Problem~\ref{prob:definition} can be regarded as a special case of the set covering optimization problem. Hence, in its naive form, it is also constrained by the facts above. 
In this work, we aim to answer the following question: Can one do better than the greedy algorithm, by utilizing the geometric structure of the objects and combining them with a learning paradigm?

\subsection{Reinforcement Learning}
The learning paradigm we use in this paper is the standard \emph{reinforcement learning} setting where an \emph{agent} learns to accomplish a certain task by interacting with an \emph{environment} over a number of discrete time steps. We restrict our attention to the approaches which are mainly built around estimating a so-called value function.

View planning can be cast as a finite \emph{Markov Decision Process} (MDP). Hence, in principle, we will be using RL techniques to solve a finite MDP. Formally, a finite MDP is a quintuple $(S,A,T,R,\gamma)$, where
$S$ denotes a finite set of Markovian states,
$A = \bigcup_{s\in S} A_s$ denotes the finite collection of all admissible actions. In particular, $A_s$ denotes the finite set of all admissible actions at state $s\in S$.
$T = \{T_a\}_{a\in A}$ is the collection of all transition probability functions. For any $(s,s') \in S\times S$, and $a\in A_s$,
$$T_a(s,s') = Pr\{s_{t+1} = s'| s_t=s \text{ and } a_t=a\}$$
is the probability that system reaches state $s'$ at time $t+1$, after taking action $a$ at state $s$. The reward signal $r_t :  S\times S \to \mathbb{R}$ returns the (expected) immediate reward received after transitioning from state $s$ to $s'$ at time $t$.
Lastly, $\gamma \in [0,1]$ is the discount factor, which simply allows us to emphasize the importance of present rewards over future ones.

In most RL systems, the state is basically agent's observation of the
environment. It can be a complete or rough estimate of the current
status of the environment. At any given state the agent chooses its
action according to a \emph{policy}. Hence, a policy is a road map for
the agent, which determines the action to take at each state. Once the
agent takes an action, the environment returns the new state and the
immediate reward. Then, the agent uses this information, together with
the discount factor to update its internal understanding of the
environment, which, in our case, is accomplished by updating a value
function.

One can use different RL algorithms to solve an MDP. In this paper we specifically use the well-known SARSA, Watkins-Q and Temporal Difference (TD) algorithms with function approximation. For a given policy $\pi$, SARSA and Watkins-Q algorithms learn $q_\pi(s,a)$, namely the \emph{action value function}, which is defined as the expected discounted total reward (i.e. return) after taking the action $a$ at state $s$ and following the policy $\pi$
\begin{equation}
q_\pi(s,a) = \mathbb{E}_\pi\{\sum_{k=0}^{\infty} \gamma^k r_{t+k+1} | s_t = s\text{ and }a_t=a\}
\end{equation}
The TD algorithm, on the other hand, learns the so-called \emph{state value function}, $v_\pi(s)$ for a state $s$. In a similar fashion, it is defined as the expected discounted total reward starting from the state $s$ and following the policy $\pi$
\begin{equation}
v_\pi(s) = \mathbb{E}_\pi\{\sum_{k=0}^{\infty} \gamma^k r_{t+k+1} | s_t = s\}
\end{equation}

\section{Reinforcement Learning for View Planning}
The simplest approach to solve VPP in RL framework would be defining each available view point at a given state as an admissible action. However, in practice, this approach would not be feasible. In our setting, the size of the state space increases exponentially with the increasing number of predefined view points. If there is no rule restricting the admissible actions the problem would quickly become intractable. In order to be able to place the problem in RL framework and solve it efficiently, one desperately needs a strategy to reduce the number of admissible actions at each state, while keeping the problem sufficiently general. 

Our inspiration in reducing the admissible actions comes from the
human approach to the problem. As we argued in
Section~\ref{sec:intro}, we claim that a human would choose non-greedy
steps in between greedy ones to solve the VPP. We model the intuitive
behavior of the human agent by using the family of functions
$f_\lambda : 2^\Omega \to \mathbb{R}$, defined as
\begin{equation}\label{eq:score_fn}
f_\lambda (X) := \frac{\mathcal{A}(X)}{\mathcal{L}(X)^\lambda}.
\end{equation}
Here $\mathcal{A}(X)$ denotes the total surface area covered by the submesh $X$, $\mathcal{L}(X)$ denotes the total boundary length of the area covered by $X$, and $\lambda \in \mathbb{R}_{\geq 0}$. 
Now, we claim that using the functions $f_\lambda$, the behavior of the human agent can be modeled as follows: 
\begin{quote}
At each step pick a $\lambda$ and choose the set which maximizes the function $f_\lambda$.
\end{quote}
As one can immediately notice, in this setting, choosing $\lambda = 0$
corresponds to proceeding greedily, whereas nonzero $\lambda$'s allow
non-greedy steps. In other words, if $\lambda\neq 0$, given two view
points introducing two different coverages $X_1$ and $X_2$ with the
same surface area, $\mathcal{A}(X_1) = \mathcal{A}(X_2)$, maximizing
$f_\lambda$ implies that the algorithm prefers the view point that
introduces a covered area with shorter perimeter (See
Algorithm~\ref{algo:NBV}). 

As one can immediately notice, in this setting, choosing $\lambda = 0$
corresponds to proceeding greedily, whereas nonzero $\lambda$'s allow
non-greedy steps. In other words, if $\lambda\neq 0$, given two view
points introducing two different coverages $X_1$ and $X_2$ with the
same surface area, $\mathcal{A}(X_1) = \mathcal{A}(X_2)$, maximizing
$f_\lambda$ implies that the algorithm prefers the view point that
introduces a covered area with shorter perimeter (See
Algorithm~\ref{algo:NBV}).

For a fixed $\lambda \geq 0$, we call the approach of maximizing $f_\lambda$ at each step, as $\lambda-$\emph{greedy algorithm}. For high $\lambda$ values, the $\lambda-$\emph{greedy algorithm} proceeds quite conservatively, preferring shorter boundaries over larger coverage, in return, causing increased number of views. Therefore, fixing $\lambda$ from the very beginning results in poor solutions. As we argued above, like a human agent does, we need to employ different values of $\lambda$
at each step. Therefore, the VPP boils down to the following decision problem:
\begin{quote}
Which $\lambda \geq 0$ to choose at each step?
\end{quote}
As we will see in the experiments section, achieving a performance better than the purely greedy approach requires a subtle choice of $\lambda$ at every step. In our experiments, we see that an ad hoc approach like alternating the $\lambda$ value between zero and a non-zero value would rarely lead to the best results. A more sophisticated strategy is needed to generate a sequence of $\lambda$'s that would lead to smaller number of views. 
\begin{algorithm}
\caption{Next Best View Selection}
\label{algo:NBV}
\begin{algorithmic}[1]
\Function{NBV}{$\lambda$}
\State $\mathcal{S} \gets 0$
\State {$\mathcal{F} \gets$ currently covered submesh}
\State {$\mathcal{E} \gets$ edges in $\mathcal{F}$}
\For{$c\ \in view\ point\ list$}
\State {$f$ $\gets$ submesh observed by $c$}
\If{$ (f \cap  \mathcal{F} \neq \emptyset) || (\mathcal{F} == \emptyset) $}
\State $s \gets$ \Call{Compute Score}{$f\cup \mathcal{F}, \lambda$} \Comment{Eq. \ref{eq:score_fn}}
\If{$s > \mathcal{S}  $}
\State $\mathcal{S} \gets s$
\State $\mathcal{C} \gets c$
\EndIf
\EndIf
\EndFor
\Return $\mathcal{C}$
\EndFunction
\end{algorithmic}
\end{algorithm}

\begin{rem}
A crucial component of our implementation is to calculate the boundary of a union of two submeshes. For two submeshes $X_1,X_2\subseteq \Omega$, we calculate the boundary $bd(X_1\cup X_2)$ according to
\begin{align}
bd(X_1\cup X_2) = &\left[bd(X_1) \setminus ed(X_2)\right] \cup \left[bd(X_2) \setminus ed(X_1)\right]\nonumber\\
											& \cup \left[bd(X_1) \cap bd(X_2)\right] \label{eq:boundary} 
\end{align}
where $ed(\cdot)$ denotes the set of all edges of the submesh. 
\end{rem}

\subsection{Beating Greedy by Learning $\lambda$}
Even though $\lambda$ is a continuous variable, we expect that the function assigning $\lambda$'s to the associated view is piece-wise continuous. Therefore, we can consider a small, finite set of $\lambda$'s to choose from at each step of our algorithm.
In order to find a sequence of $\lambda$'s that leads to a solution better than the one offered by the greedy algorithm, we device a RL scheme. In this setup, our state is a vector of length equals to the number of initial view points, which is denoted by $N$. The set of chosen view points uniquely define the state: If at a given state, the view point $i$ is chosen, then the $i^{th}$ entry of the state vector is set, otherwise it remains zero. This way, we introduce a state space with $2^N$ states. Obviously, this definition of the state satisfies the Markov property. In this setting, at each state, taking an action corresponds to choosing a $\lambda$ value. However, the learning agent is allowed to choose a $\lambda$ value only from a finite set of admissible $\lambda$'s, which is denoted by $\Lambda$. We assume that $\Lambda$ remains unchanged at each state. We further assume that the agent follows a deterministic policy, hence all transition probabilities are trivial. Since we would like to accomplish the coverage in as few steps as possible, we introduce a reward of $-1$ for each state transition. We don't use any discount factor, and the coverage task is naturally episodic.
In this setting, the VPP becomes a finite Markov Decision Process. 

\subsection*{Learning stage}
In order to solve this MDP, we use three different RL algorithms: On-policy control algorithm SARSA, off-policy control algorithm Watkins-Q and on-policy learning algorithm TD. The former two algorithms learn $q_\pi(s,a)$, the action value function, whereas the last algorithm learns $v_\pi(s)$, the state value function. 
\begin{algorithm}[t]
\caption{Watkins-Q Agent}
\label{algo:WatkinsQ}
\begin{algorithmic}[1]
\Procedure{Learning}{} 
\State{$ \theta \gets  random\ network\ weights$}
\State{$\alpha \gets learning\ rate$, $\mu_e \gets eligibility\ factor$}
\State{$\varepsilon \gets exploration\ probability$}
\Repeat
\State {$c \gets$ random view point}
\State{$ s \gets \{c\}$, $e \gets 0$, $ r \gets -1$, $\delta \gets 0$}
\If{$random\ number > \varepsilon$}
\State $\lambda^{*} \gets \text{arg}\max\limits_{\lambda}\hat{q}_\pi(\theta, s, \lambda)$
\Else
\State $\lambda^* \gets random\ \lambda\ from\ \Lambda$
\EndIf
\While{$true$}
\State $e \gets e + \nabla_{\theta}\hat{q}_\pi(\theta, s, \lambda^{*})$
\State $\delta  \gets r - \hat{q}_\pi(\theta, s, \lambda^{*})$
\If{$ s\ is\ \emph{Terminal}$}
\State $\theta \gets \theta + \alpha \cdot \delta \cdot e$
\State $break$
\EndIf
\State $c \gets$ \Call{NBV}{$\lambda^{*}$}\Comment{see Alg. \ref{algo:NBV}}
\State $ s \gets s \cup \{c\}$
\State $\delta \gets \delta + \max\limits_{\lambda}\hat{q}_\pi(\theta, s, \lambda))$
\State $\theta \gets \theta + \alpha \cdot \delta \cdot e$
\If{$random\ number > \varepsilon$}
\State $\lambda^{*}\! \gets\! \text{arg}\max\limits_{\lambda}\hat{q}_\pi(\theta, s, \lambda)$, $e \gets \mu_{e} \cdot e$
\Else
\State $\lambda^* \gets random\ \lambda\ from\ \Lambda$, $e \gets 0$
\EndIf
\EndWhile
\Until{Max nr of episodes is reached}
\EndProcedure
\end{algorithmic}
\end{algorithm}
\begin{algorithm}[t]
\caption{TD Agent}
\label{algo:TD}
\begin{algorithmic}[1]
\Procedure{Learning}{} 
\State{$ \theta \gets  random\ network\ weights$}
\State{$\alpha \gets learning\ rate$, $\mu_e \gets eligibility\ factor$}
\Repeat
\State {$c \gets$ random view point}
\State{$ s \gets \{c\}$, $e \gets 0$, $ r \gets -1$, $\delta \gets 0, \mathcal{S}\gets \emptyset$}
\While{$true$}
\State $e \gets e + \nabla_{\theta}\hat{v}_\pi(\theta, s)$
\State $\delta  \gets r - \hat{v}_\pi(\theta, s)$
\If{$ s\ is\ \emph{Terminal}$}
\State $\theta \gets \theta + \alpha \cdot \delta \cdot e$
\State $break$
\EndIf
\State $\mathcal{S} \gets \{s\cup\{c\} | c =  \Call{NBV}{\lambda}\ and\ \lambda\in\Lambda\}$
\State $s \gets \text{arg}\max\limits_{s'\in \mathcal{S}} \hat{v}_\pi(\theta, s')$
\State $\delta \gets \delta + \hat{v}_\pi(\theta, s)$
\State $\theta \gets \theta + \alpha \cdot \delta \cdot e$
\State $e \gets \mu_{e} \cdot e$
\EndWhile
\Until{Max nr of episodes is reached}
\EndProcedure
\end{algorithmic}
\end{algorithm}
For the convenience of the reader, we include our learning procedures implementing Watkins-Q and TD algorithms in the algorithm boxes \ref{algo:WatkinsQ} and \ref{algo:TD} (We refer to the supplementary material for the implementation of SARSA). In these algorithms, we call a state terminal if a certain coverage criteria is met. Our coverage criteria is relative in the sense that the coverage task is assumed to be completed once we cover a certain percentage of the area that can be covered by the union of all initial view points. We call this number the \emph{relative coverage criteria}, or \emph{RCC}. In order to boost the learning performance, we use eligibility traces. Moreover, since the number of states quickly becomes huge, we need to deploy a function approximation scheme. In all cases the value function is approximated by a neural network with one hidden layer which has sigmoid neurons. The output layer of the neural network is an affine function, i.e. a linear function with weights and a bias term. In case of SARSA and Watkins-Q, the input to the network is the concatenation of the state vector and a one-hot action vector encoding the chosen $\lambda$. In order to achieve this, we basically enumerate the admissible $\lambda$'s in $\Lambda$ and the $i^{th}$ entry of the action vector is set if the corresponding $\lambda$ is chosen. As for the implementation of TD with function approximation, the input to the network is the state vector only. 

\begin{figure}
\centering
\includegraphics[width=0.56\linewidth, trim = 0 0 0 0, clip]{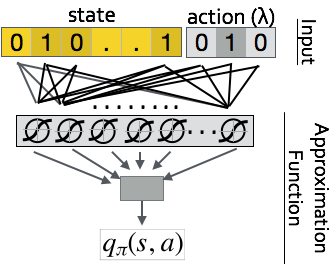}
\caption{Function approximation using neural network in SARSA and Watkins-Q settings. Note that, in TD method, actions are omitted from the input.}
\label{fig:nnet}
\end{figure}
We refer to Figure \ref{fig:nnet} for an illustration of the value function network. If we let $\sigma_i$ denote the output of the $i^{th}$ hidden sigmoid neuron, $w_i$ and $b$ denote the weights and the bias of the output layer, then the output of the neural network, $\Phi$, is given by
\begin{equation}
\Phi = b + \sum w_i\sigma_i
\end{equation}
Furthermore, if we let $w_{ij}$ denote the weights and $b_i$ denote the bias of the $i^{th}$ hidden sigmoid neuron, then the gradient of the network, which is required for the implementation of the algorithms above, can be calculated by
\begin{subequations}
\begin{alignat}{3}
\frac{\partial\Phi}{\partial b} &= 1, \quad\quad\frac{\partial\Phi}{\partial b_i} &&= w_i\sigma_i(1-\sigma_i),\\
\frac{\partial\Phi}{\partial w_i} &= \sigma_i,\quad \frac{\partial\Phi}{\partial w_{ij}} &&= w_i\sigma_i(1-\sigma_i)x_j
\end{alignat}
\end{subequations}

\subsection*{Planning using the policy $\pi$}\label{ssect:Prediction}
Once we build a system that estimates the action or state values, a policy which will suggest a solution to the VPP can be derived quite easily. The derived policy acts greedily with respect to the estimated values. To be more precise, in case of SARSA and Watkins-Q, at each state, we go through all admissible actions, find the action which has the highest value, and take that action to move to the next state, until the coverage task is completed. On the other hand, in case we are using the TD algorithm, at each state we calculate all possible next states by going through all admissible actions and finally pick the action that leads to the state with the highest estimated value. This process eventually produces a sequence of $\lambda$'s that solves the VPP.

\section{Experimental Setup}
In order to test the performance of the solution method we proposed, we experimented on $3D$ meshes of $20$ different objects. The first $8$ objects consisted mostly of those which could be of potential interest in the application areas we mentioned in the introduction. Particularly, we tested the method on the $3D$ model of a mountainous region, Yosemite Valley, a wind turbine, a skull, Statue of Liberty, an engine block, and finally, as toy examples, a knot and a plane. The second group of test objects are obtained from the data set appeared in \cite{Hint:13}. For each model, we tested our method against two different methods: Purely Greedy and Alternating-$\lambda$. As the name suggests, in Purely Greedy approach, we basically complete coverage by proceeding greedily at each step, whereas in Alternating-$\lambda$ case, after starting with a greedy step, we let $\lambda$ alternate between $0$ and $1$ sequentially, and choose the view which maximizes the score function \eqref{eq:score_fn} at each step. We used a virtual rgb camera as a sensor, and for fair comparison of the methods, for each object we kept the initial set of cameras and their settings fixed while changing the solution method. 
\begin{figure}[!t]
\centering
\setlength\tabcolsep{0.075cm}
\begin{tabular}{cccc}
	Y. Valley & Wind Turb. & Stat. Liberty & Engine\\
     	\includegraphics[height=.17037\linewidth, width=.23\linewidth]{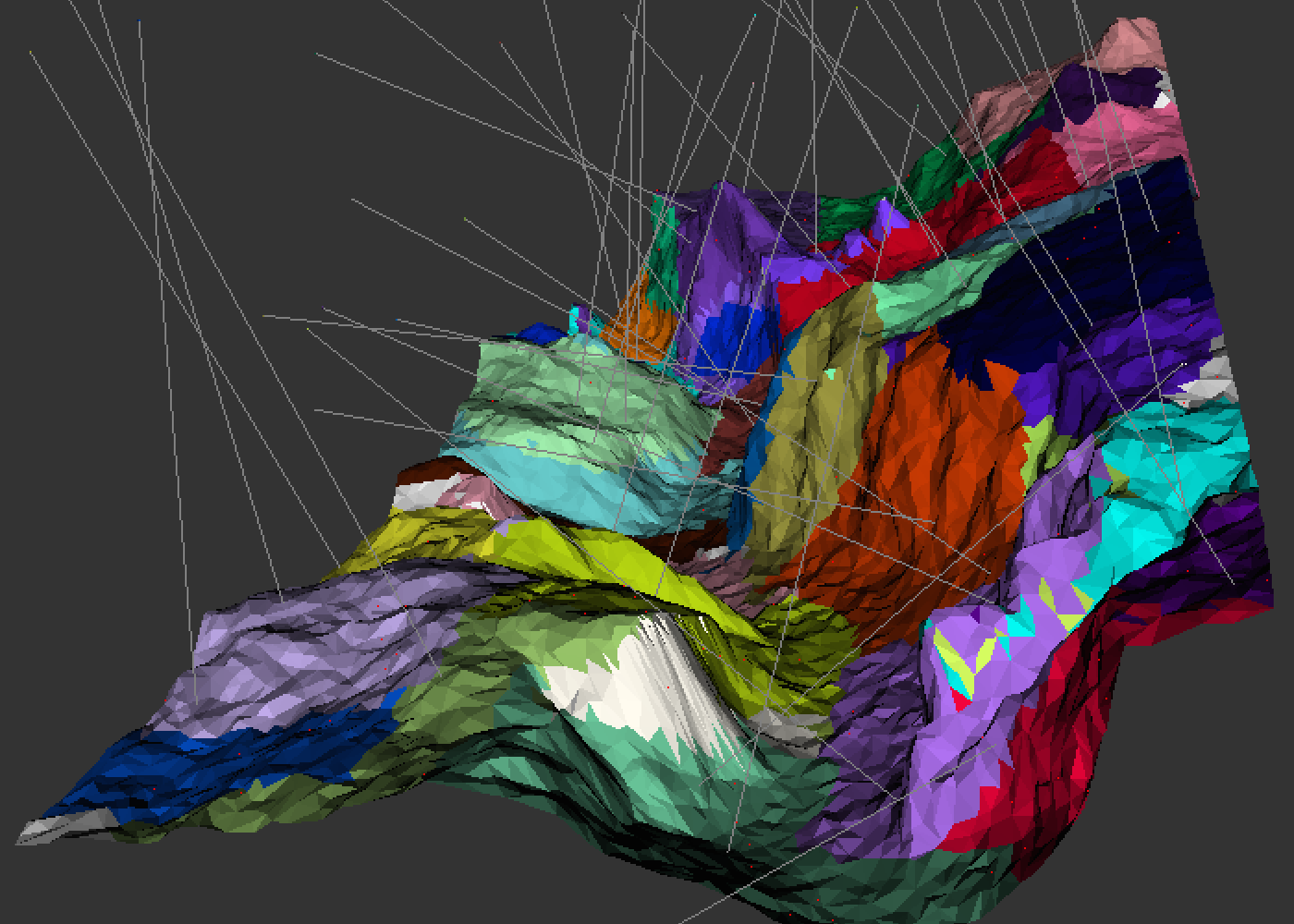}&
    	\includegraphics[height=.17037\linewidth, width=.23\linewidth]{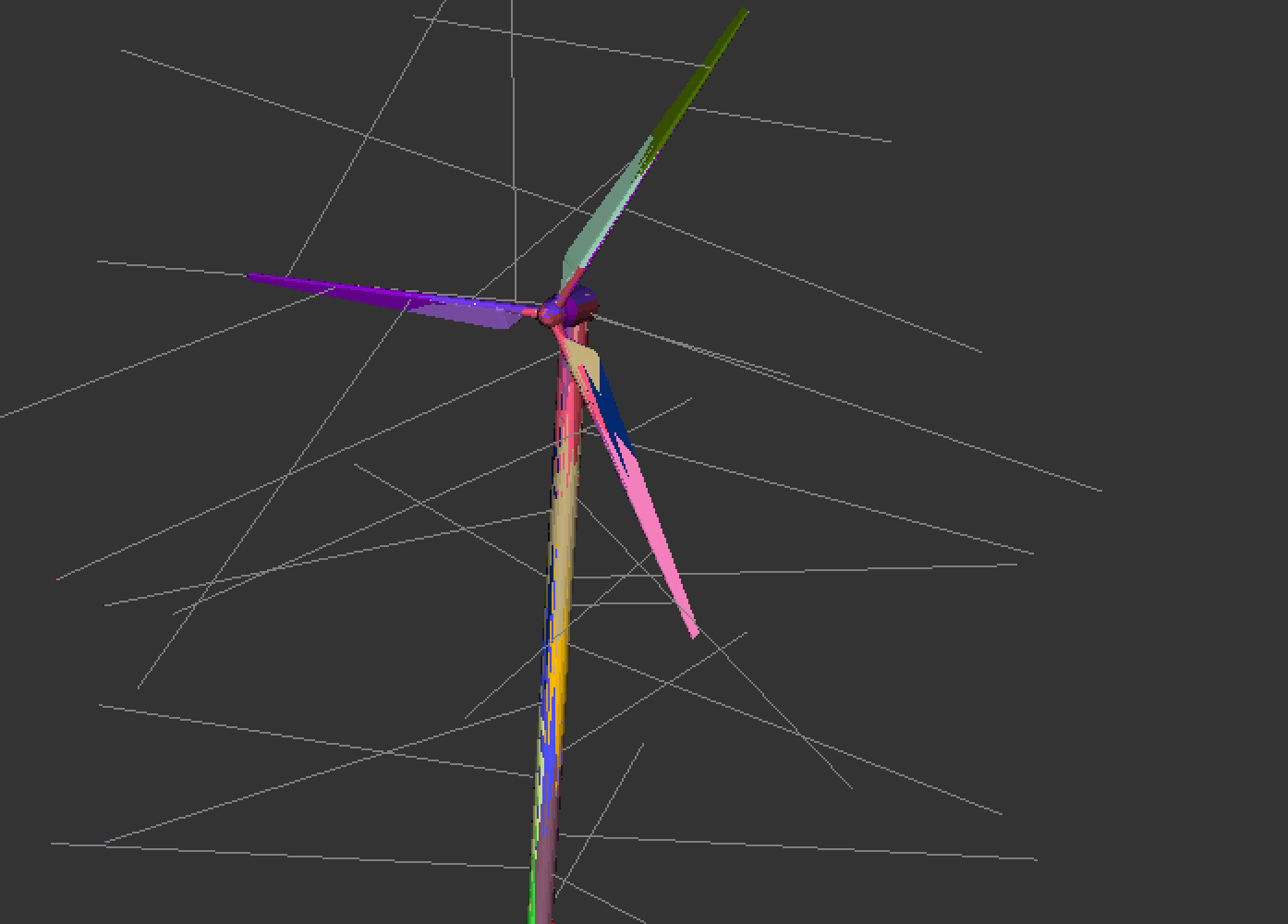} &
    	\includegraphics[height=.17037\linewidth, width=.23\linewidth]{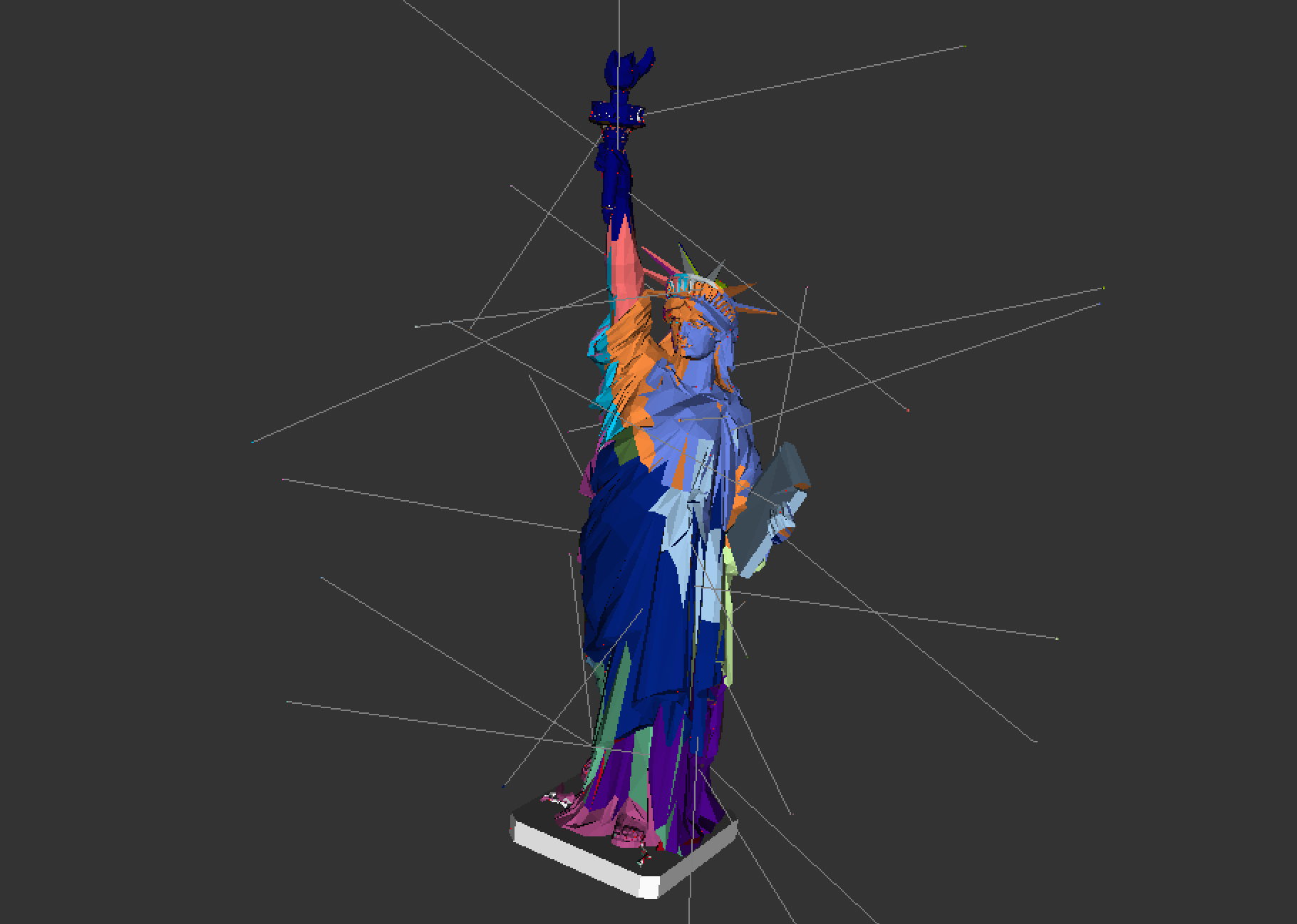} &
        	\includegraphics[height=.17037\linewidth, width=.23\linewidth]{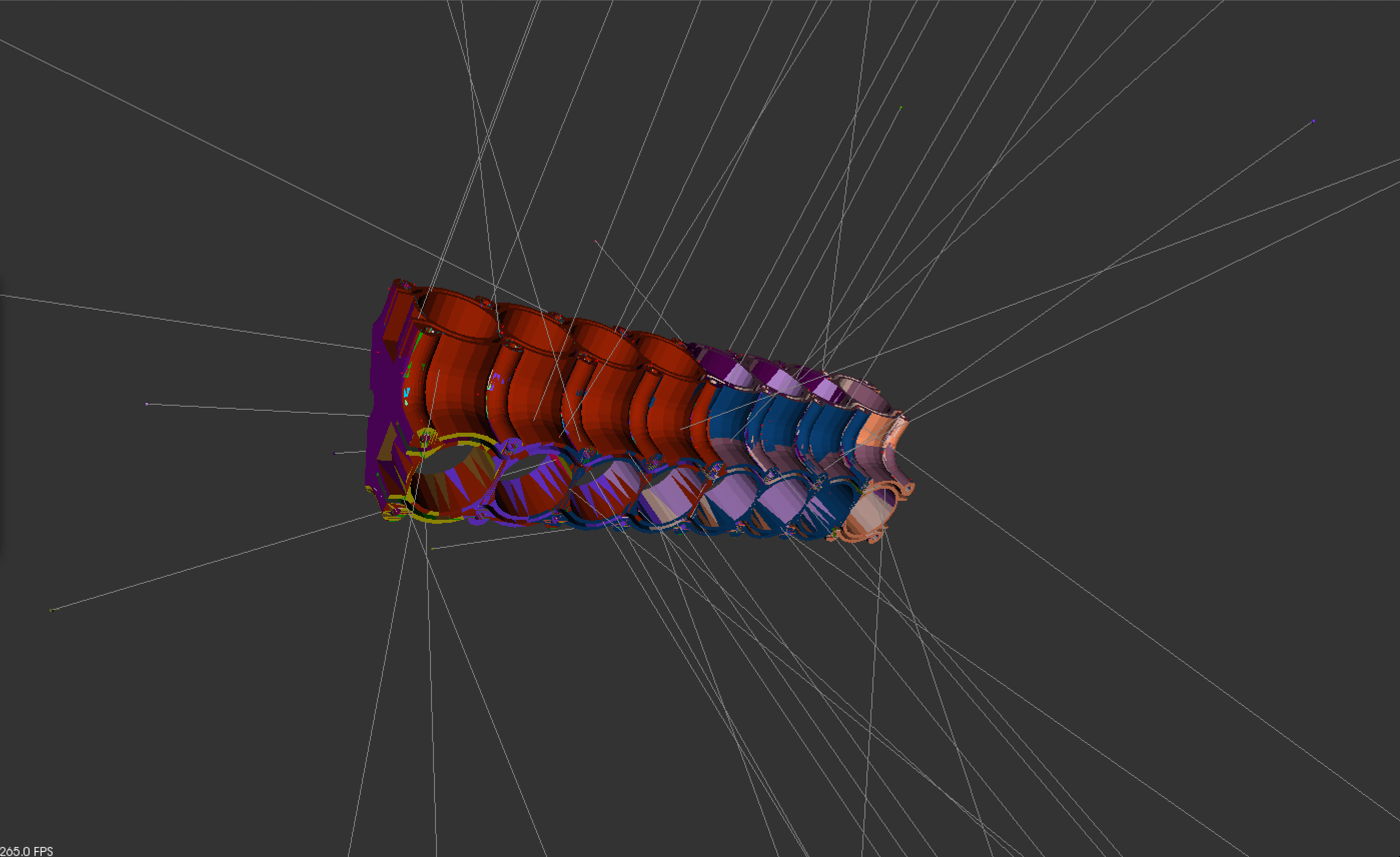}  \\
	\includegraphics[height=.3123\linewidth, width=.23\linewidth]{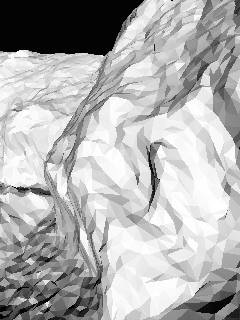} &
	\includegraphics[height=.3123\linewidth, width=.23\linewidth]{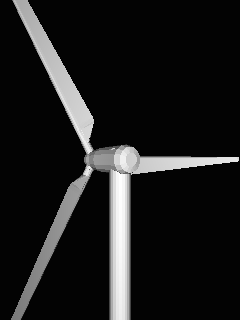} &
	\includegraphics[height=.3123\linewidth, width=.23\linewidth]{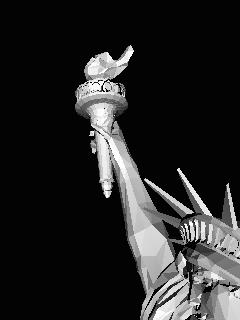}&
	\includegraphics[height=.3123\linewidth, width=.23\linewidth]{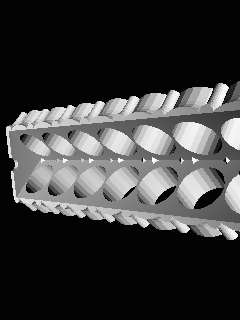} \\
\end{tabular}
\caption{Visual results of coverage and sample views on various models. In the top row, lines represent location and direction of the selected cameras. Colors represent coverage by different cameras. Best seen in color and electronic format.}
\label{fig:data}
\end{figure} 
Figure~\ref{fig:data} shows a few of the models (with coverage map) from the first group and their sample views. The images of the rest of the models can be found in the supplementary material.

As we mentioned previously, we used three different reinforcement learning algorithms to implement our method. During these implementations we allowed only two actions: $\lambda = 0$ and  $\lambda = 1$. The hidden layer of the value network included $200$ neurons. We used eligibility traces with eligibility factor equals $0.5$. The learning rate was set to $0.01$ and the maximum number of episodes was set to $100K$. Once this maximum number is reached, we terminated the learning phase, and ran the trained network to accomplish the coverage task. For the first group of objects we mentioned above, RCC was set to $0.99$, for the second set this number was increased to $1.0$. During the learning and the planning phases same RCC was targeted. In the learning phase of the Watkins-Q agent, we allowed exploration within the first 50K episodes, whereas for SARSA and TD agents no exploration was allowed.

\section{Results and Discussions}
Given sufficient time for learning and exploration, our method is expected to perform at least as good as the Purely Greedy or Alternating-$\lambda$ approaches.
\begin{table*}[!ht]
\label{tb:results}
\begin{center}
\setlength\tabcolsep{0.08cm}
\begin{tabular}{|l|c|c|c|c|c|c|c|c|c|a|a|a|a|a|a|a|a|a|a|a|a||cc|cc|}
\hline
\multicolumn{2}{|c|}{~} & \rotatebox[origin=c]{90}{Mountains} & \rotatebox[origin=c]{90}{Valley} & \rotatebox[origin=c]{90}{Turbine} & \rotatebox[origin=c]{90}{Skull} & \rotatebox[origin=c]{90}{Statue} & \rotatebox[origin=c]{90}{Engine} & \rotatebox[origin=c]{90}{Knot} & \rotatebox[origin=c]{90}{Plane} & \rotatebox[origin=c]{90}{Ape} & \rotatebox[origin=c]{90}{Cat} & \rotatebox[origin=c]{90}{Iron} & \rotatebox[origin=c]{90}{Can} & \rotatebox[origin=c]{90}{Lamp} & \rotatebox[origin=c]{90}{Phone} & \rotatebox[origin=c]{90}{Glue} & \rotatebox[origin=c]{90}{Driller} & \rotatebox[origin=c]{90}{Eggbox} & \rotatebox[origin=c]{90}{Cam} & \rotatebox[origin=c]{90}{Duck} & \rotatebox[origin=c]{90}{Bench} & \rotatebox[origin=c]{90}{\footnotesize Avg. Learning} & \rotatebox[origin=c]{90}{\footnotesize Time (Ep/sec)} & \rotatebox[origin=c]{90}{\footnotesize Avg. Planning} & \rotatebox[origin=c]{90}{\footnotesize Time (Ep/sec)}  \\ \hline
\multicolumn{2}{|l|}{\# of init. cams} &$376$&$270$&$264$&$270$&$265$&$289$&$286$&$189$&$312$&$332$&$333$&$412$&$245$&$302$&$344$&$342$&$319$&$342$&$343$&$321$ &\multicolumn{2}{|c|}{n/a} &\multicolumn{2}{|c|}{n/a} \\ \hline
\multicolumn{2}{|l|}{Greedy}                 	& $36$ 	    & $42$ 		& $\tcr{34}$  & $39$ 		& $31$ 	   & $50$ 		& $\tcr{13}$  & $16$ 		& $\tcr{22}$ & $\tcr{15}$ & $16$ 	  & $\tcr{16}$  & $26$ 	 & $13$ 	     & $\tcr{13}$ 	& $16$ 	    & $11$ 		& $11$ 	  & $\tcr{9}$   & $\tcr{30}$ &\multicolumn{2}{|c|}{n/a}  &\multicolumn{2}{|c|}{$0.17$}\\ \hline
\multicolumn{2}{|l|}{Altern. $\lambda$} 	& $\tcr{38}$  & $\tcr{43}$ 	& $33$ 	    & $\tcr{42}$ 	& $\tcr{32}$ & $\tcr{52}$ 	& $12$ 	    & $\tcr{17}$  & $19$ 	   & $12$ 	      & $\tcr{17}$   & $\tcr{16}$ & $\tcr{28}$  & $\tcr{14}$  & $12$ 	& $\tcr{19}$  & $\tcr{13}$ 	& $\tcr{12}$& $\tcb{7}$  & $29$ &\multicolumn{2}{|c|}{n/a} &\multicolumn{2}{|c|}{$0.17$} \\ \hline \hline
\multirow{3}{*}{\rotatebox[origin=c]{90}{Ours}} 
&SARSA 			& $\tcb{34}$ & $\tcb{39}$  & $\tcb{32}$ & $\tcb{37}$ & $\tcb{29}$ & $\tcb{48}$ 	& $\tcb{11}$ & $\tcb{13}$ & $\tcb{17}$ & $\tcb{11}$ & $15$ 	  & $\tcb{13}$ & $\tcb{23}$ & $\tcb{11}$ & $\tcb{10}$ & $\tcb{15}$ & $11$ 		& $\tcb{9}$ & $8$ 	     & $\tcb{26}$  &\multicolumn{2}{|c|}{$0.52$} &\multicolumn{2}{|c|}{$0.51$} \\ \cline{2-26}
&Watkins-Q 		& $\tcb{34}$ & $\tcb{39}$  & $\tcb{32}$ & $\tcb{37}$ & $\tcb{29}$ & $\tcb{48}$ 	& $\tcb{11}$ & $\tcb{13}$ & $\tcb{17}$ & $\tcb{11}$ & $\tcb{14}$ & $\tcb{13}$ & $\tcb{23}$ & $\tcb{11}$ & $\tcb{10}$ & $\tcb{15}$ & $11$		& $\tcb{9}$ & $8$  	     & $\tcb{26}$  &\multicolumn{2}{|c|}{$0.51$} &\multicolumn{2}{|c|}{$0.50$} \\ \cline{2-26}
&TD 				& $\tcb{34}$ & $\tcb{39}$  & $\tcb{32}$ & $\tcb{37}$ & $\tcb{29}$ & $\tcb{48}$ 	& $\tcb{11}$ & $\tcb{13}$ & $\tcb{17}$ & $\tcb{11}$ & $\tcb{14}$ & $\tcb{13}$ & $\tcb{23}$ & $\tcb{11}$ & $\tcb{10}$ 	& $\tcb{15}$ & $\tcb{10}$  & $\tcb{9}$ & $8$ 	     & $\tcb{26}$  &\multicolumn{2}{|c|}{$1.12$} &\multicolumn{2}{|c|}{$1.01$} \\ \hline
\end{tabular}
\end{center}
\caption{Comparison of the performance of different algorithms on different $3D$ models. Columns show the number of cameras proposed by each method. Last column shows the duration of a single episode.}
\label{tb:results}
\end{table*}
As expected, we see from Table~\ref{tb:results} that in almost all test cases, our method provides a solution which is better than the solution provided by either of the baseline methods. An exception to this is the \emph{duck} data, where the Alternating-$\lambda$ approach performed surprisingly better than any other method. However, in general, even without introducing explicit exploration, our reinforcement learning based method successfully reduces the number of cameras required to ensure the coverage of the object. 
\begin{figure}[b]
\setlength\tabcolsep{0.0cm}
\centering
\rotatebox[origin=c]{90}{\footnotesize Avg. \# of cameras} 
\begin{tabular}{c}
    \includegraphics[width=0.9\linewidth, trim = 0 0 0 0, clip]{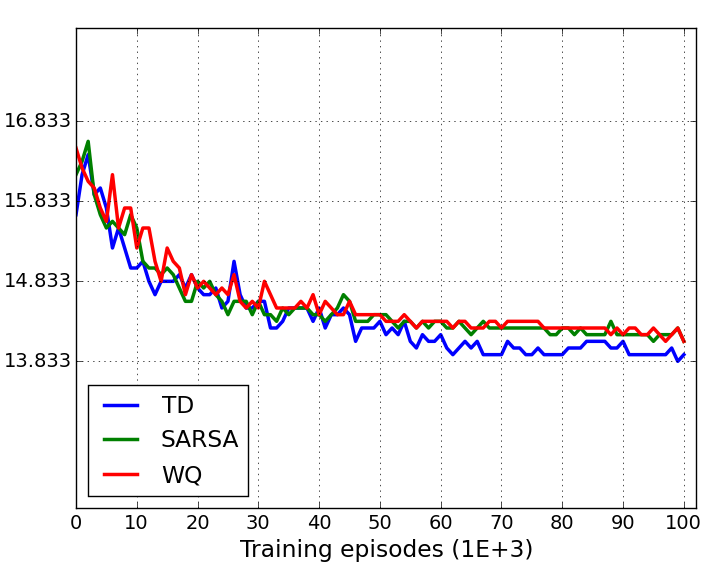}
\end{tabular}
\caption{Average convergence performance of RL algorithms using twelve models from \cite{Hint:13}.}
\label{fig:Convergence}
\end{figure}
In this set of experiment, we limited the learning phase to $100K$ episodes and as shown in Figure~\ref{fig:Convergence}, we observed that the average performance of agents did not change significantly after $65K$ episodes.

Another interesting result reflected by this experiment is that, when the RCC is not $1.0$, the adverse effects of the purely greedy approach is less visible. This is simply because when RCC is $0.99$, a solution leaving small uncovered areas behind is considered a success, even though there cameras in the initial view point set seeing those uncovered areas. This relaxation works very much in favor of the purely greedy algorithm, which already tends to leave plenty of those small uncovered areas while maximizing the overall coverage. Note that, as mentioned before, the RCC was $0.99$ for the first set of $8$ objects, and $1.0$ for the remaining $12$ in the table and we see that the average performance gain of RL based methods over the purely greedy approach in the second set, is shrunk from $4$ to $2$ view points for the first set of objects.

\begin{figure*}[t!]
\centering
\begin{minipage}{1\linewidth}
\rotatebox[origin=c]{90}{\footnotesize Avg. \# of Cams}
\setlength\tabcolsep{-0.19cm}
\hspace{0.01in}
\begin{tabular}{ccccc}
	\centering
	RCC = $1.0$ & RCC = $0.98$ & RCC = $0.96$ & RCC = $0.94$ & RCC = $0.92$ \\ 
	\includegraphics[width=.218\linewidth, trim=40 20 0 30, clip ]{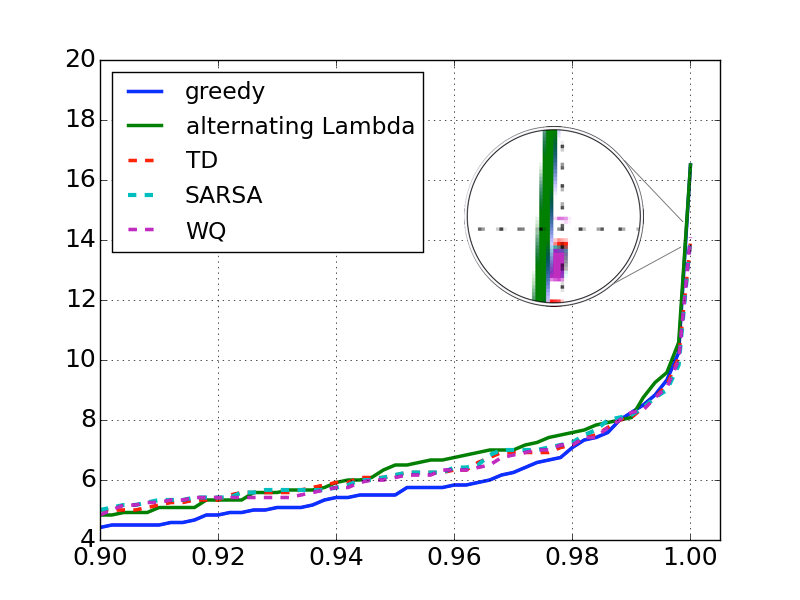} &
	\includegraphics[width=.218\linewidth, trim=40 20 0 30, clip ]{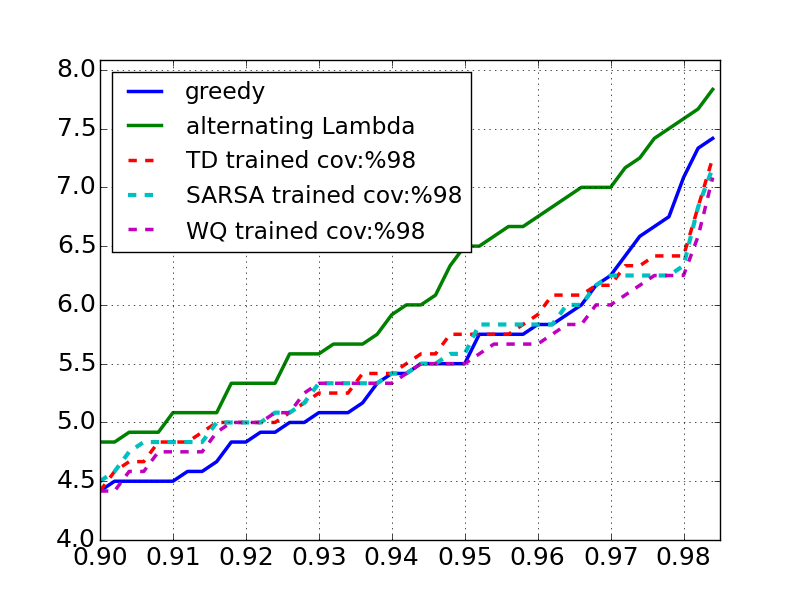} &
	\includegraphics[width=.218\linewidth, trim=40 20 0 30, clip ]{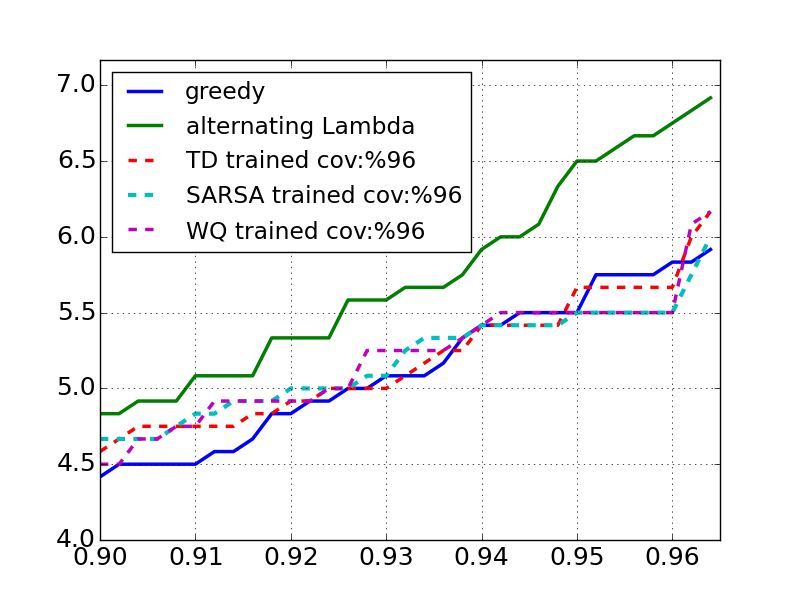} &
	\includegraphics[width=.218\linewidth, trim=40 20 0 30, clip ]{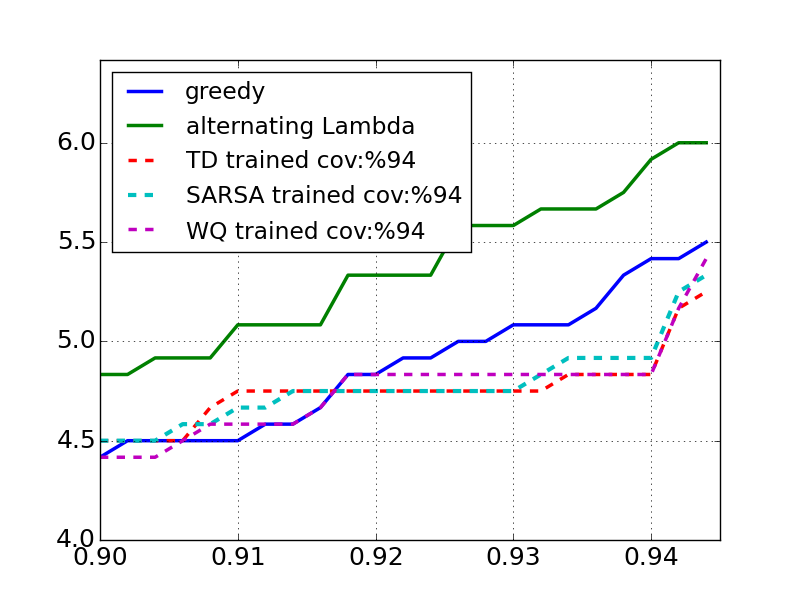}&
	\includegraphics[width=.218\linewidth, trim=40 20 0 30, clip ]{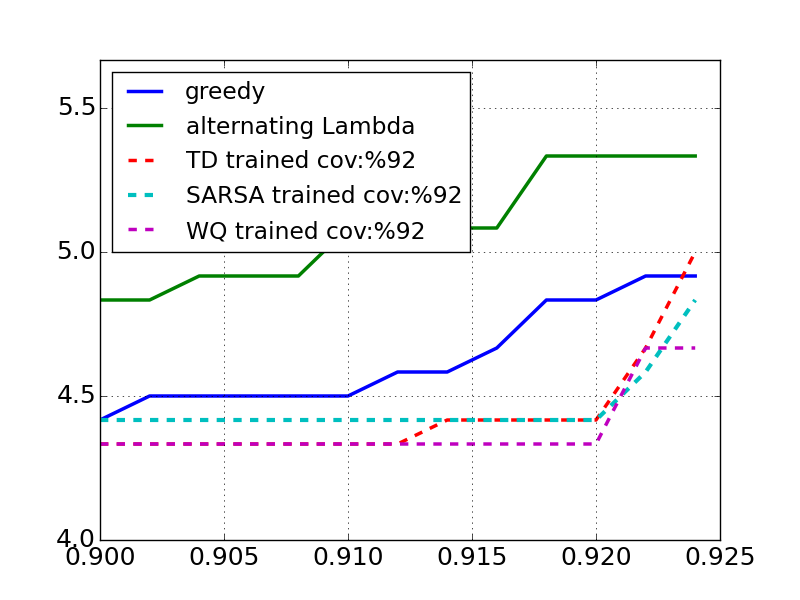} \\
	& &\footnotesize Relative Coverage & & 
\end{tabular}
\end{minipage}
\caption{Comparison of the average performance of different algorithms with varying relative coverage criteria (RCC). The average is taken over the second dataset, which consists of $12$ models.}
\label{fig:rpct_sweep}
\end{figure*}

For a thorough analysis of RCC on the performance of our method, in an auxiliary experiment we retrained all three RL-based systems for different RCC values. The results of this auxiliary experiment is summarized in Figure~\ref{fig:rpct_sweep}. Each plot in this figure compares the average performance of baseline methods against the average performance of RL-based methods. The performance average is obtained after running each of these algorithms for each of the $12$ objects appearing in the second data set. 
After learning, during test time we recorded the average number of cameras selected by each method when RCC is varying from $0.9$ to $1.0$. We observed appealing results: i) RL based methods beat greedy algorithms with larger margins when the coverage task is completed, i.e. RCC is met;  ii) RL agents trained with a certain RCC value can perform worse than greedy methods for lower RCC at test; iii) when RCC is set lower in both learning and planning stages, the performance gain of our RL based methods is reduced but they still perform better than greedy methods.  

\begin{figure}[t!]
\begin{minipage}{1\linewidth}
\rotatebox[origin=c]{90}{\footnotesize Estimated Returns}
\setlength\tabcolsep{-0.01cm}
\hspace{-0.05in}
\begin{tabular}{cc}
	\centering
         SARSA & TD\\ 
	\includegraphics[width=.48\linewidth]{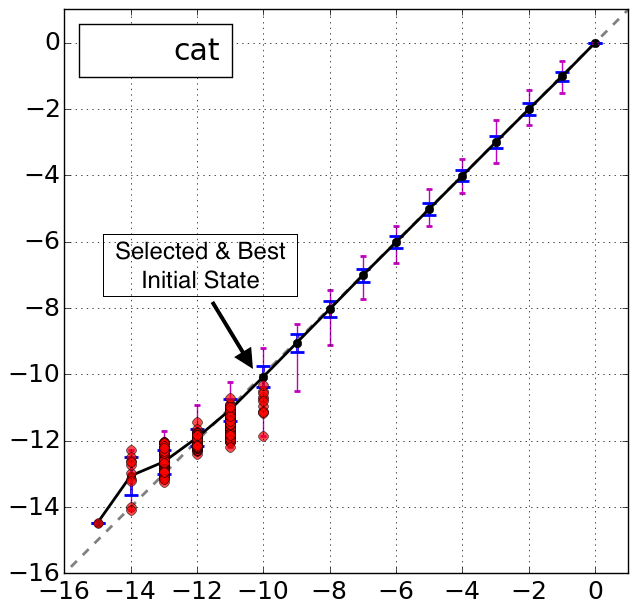} &
	\includegraphics[width=.48\linewidth]{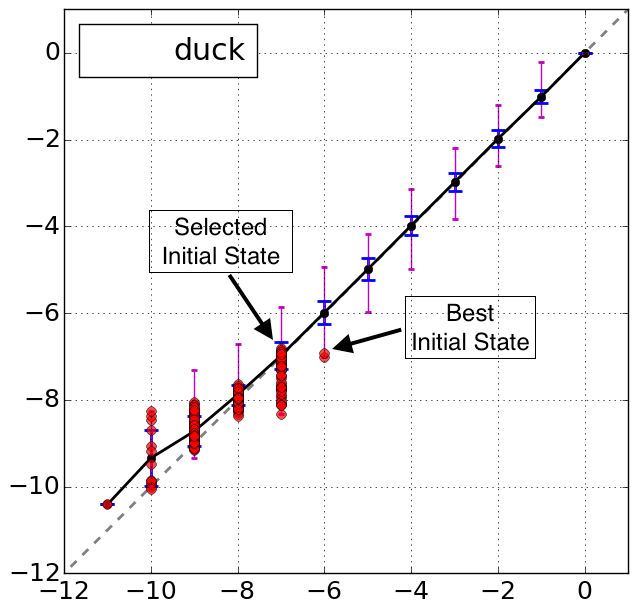} \\ 
\end{tabular}
\end{minipage}
\centering
	\footnotesize Actual Returns
\caption{Error plots for two cases. The solid black line indicates the mean, the error bars indicate the standard deviation, data min and data max. Red dots represent the initial states. We mark the actual best initial state and the initial state selected by the policy.}
\label{fig:func_approximation}
\end{figure}

Finally, in order to verify the precision of the value function approximation, we compared the actual return (i.e. the sum of actual rewards observed following the policy) and the estimated return (i.e. estimated state value in case of TD, and maximum estimated action value in case of SARSA and Watkins-Q) of a number of states. In order to do that, we collected data by starting from all possible initial states, i.e. states corresponding to a single camera only, and following the policy suggested by the network, as explained in Section~\ref{ssect:Prediction}. For each state visited, we calculated the estimated and actual returns. As a small sample of this analysis, in Figure \ref{fig:func_approximation} we include the results from experiments of \emph{duck} and \emph{cat} objects. The analysis for other objects can be found in the supplementary material. Considering the results shown in Table \ref{tb:results}, we chose two object-method pairs.
Accordingly, \emph{cat}-SARSA experiment shows an example of good approximation, and \emph{duck}-TD experiment illustrates a bad approximation. 

In the plots of Figure \ref{fig:func_approximation}, the absolute value of the actual return tells us how many cameras more we need to place in order to accomplish the coverage task. In the ideal case the estimated return and the actual return should be equal and we should see a distribution on the $y=x$ line only. We see that, in both cases the expected value of the estimated returns satisfy this property. Moreover, even though the outliers do exist, the standard deviation of the approximation error is rather small. As expected, for the states that are visited more frequently, networks provided quite good approximations, whereas the values of the states that are visited less often, e.g. the initial states (represented by red dots in Figure \ref{fig:func_approximation}), are often approximated rather poorly. However, the overall approximation quality of the networks are quite high.

This analysis helps us understand why TD method for \emph{duck} object failed to perform as good as Alternating-$\lambda$. As shown in the corresponding plot, the initial state selected by the policy and the initial state which leads to the best result differs. This is due to bad estimation of the state value of the true best initial state. 

\section{Conclusion}
In this paper, we proposed a fully automated reinforcement learning (RL) based method to solve the view planning problem (VPP) for coverage of $3D$ object models. The solution given in this paper is neither limited to structure of the $3D$ models nor the type of the sensors that are used. Given sufficient exploration and learning time, the proposed method is guaranteed to perform at least as good as the greedy algorithm. In an extensive set of test cases, we showed that our proposed method out-performs the greedy algorithm, and we further showed that a similar performance metrics cannot be attained by ad hoc approaches like Alternating-$\lambda$. A natural extension of our work is to add path planning to proposed approach and provide an extensive treatment of model-based VPP.

\clearpage
{\small
\bibliographystyle{ieee}
\bibliography{sp_sarsa_ref}

\begin{thebibliography}{10}\itemsep=-1pt

\bibitem{BlAl:07}
P.~S. Blaer and P.~K. Allen.
\newblock Data acquisition and view planning for 3-d modeling tasks.
\newblock In {\em IEEE/RSJ International Conference on Intelligent Robots and
  Systems}, pages 417--422, 2007.

\bibitem{blaasjo2005isoperimetric}
V.~Bl{\aa}sj{\"o}.
\newblock The isoperimetric problem.
\newblock {\em The American Mathematical Monthly}, 112(6):526--566, 2005.

\bibitem{RaMe:15}
F.-M. De~Rainville, J.-P. Mercier, C.~Gagn{\'e}, P.~Giguere, and D.~Laurendeau.
\newblock Multisensor placement in 3d environments via visibility estimation
  and derivative-free optimization.
\newblock In {\em 2015 IEEE International Conference on Robotics and Automation
  (ICRA)}, pages 3327--3334, 2015.

\bibitem{dinur2014analytical}
I.~Dinur and D.~Steurer.
\newblock Analytical approach to parallel repetition.
\newblock In {\em Proceedings of the 46th Annual ACM Symposium on Theory of
  Computing}, pages 624--633, 2014.

\bibitem{englot2011planning}
B.~Englot and F.~Hover.
\newblock Planning complex inspection tasks using redundant roadmaps.
\newblock In {\em Proc. Int. Symp. Robotics Research}, 2011.

\bibitem{Erdem:2006db}
U.~M. Erdem and S.~Sclaroff.
\newblock {Automated camera layout to satisfy task-specific and floor
  plan-specific coverage requirements}.
\newblock {\em Computer Vision and Image Understanding}, 103(3):156--169, 2006.

\bibitem{Fe:98}
U.~Feige.
\newblock A threshold of ln n for approximating set cover.
\newblock {\em J. ACM}, 45(4):634--652, July 1998.

\bibitem{Haque_2016_CVPR}
A.~Haque, A.~Alahi, and L.~Fei-Fei.
\newblock Recurrent attention models for depth-based person identification.
\newblock In {\em The IEEE Conference on Computer Vision and Pattern
  Recognition (CVPR)}, June 2016.

\bibitem{Hint:13}
S.~Hinterstoisser, V.~Lepetit, S.~Ilic, S.~Holzer, G.~Bradski, K.~Konolige, and
  N.~Navab.
\newblock Model based training, detection and pose estimation of texture-less
  3d objects in heavily cluttered scenes.
\newblock In {\em 11th Asian Conference on Computer Vision}. Springer Berlin
  Heidelberg, 2013.

\bibitem{lin2005near}
F.~Y. Lin and P.-L. Chiu.
\newblock A near-optimal sensor placement algorithm to achieve complete
  coverage-discrimination in sensor networks.
\newblock {\em IEEE Communications Letters}, 9(1):43--45, 2005.

\bibitem{Ma:99}
E.~Marchand and F.~Chaumette.
\newblock Active vision for complete scene reconstruction and exploration.
\newblock {\em IEEE Transactions on Pattern Analysis and Machine Intelligence},
  21(1):65--72, 1999.

\bibitem{martin2015evolutionary}
R.~A. Martin, I.~Rojas, K.~Franke, and J.~D. Hedengren.
\newblock Evolutionary view planning for optimized uav terrain modeling in a
  simulated environment.
\newblock {\em Remote Sensing}, 8(1):26, 2015.

\bibitem{Mathe_2016_CVPR}
S.~Mathe, A.~Pirinen, and C.~Sminchisescu.
\newblock Reinforcement learning for visual object detection.
\newblock In {\em The IEEE Conference on Computer Vision and Pattern
  Recognition (CVPR)}, June 2016.

\bibitem{Mi:08}
A.~Mittal and L.~S. Davis.
\newblock A general method for sensor planning in multi-sensor systems:
  Extension to random occlusion.
\newblock {\em International Journal of Computer Vision}, 76(1):31--52, 2008.

\bibitem{MnHe:14}
V.~Mnih, N.~Heess, A.~Graves, et~al.
\newblock Recurrent models of visual attention.
\newblock In {\em Advances in Neural Information Processing Systems}, pages
  2204--2212, 2014.

\bibitem{MnKa:13}
V.~Mnih, K.~Kavukcuoglu, D.~Silver, A.~Graves, I.~Antonoglou, D.~Wierstra, and
  M.~Riedmiller.
\newblock Playing atari with deep reinforcement learning.
\newblock {\em arXiv preprint arXiv:1312.5602}, 2013.

\bibitem{MoRu:16}
C.~Mostegel, M.~Rumpler, F.~Fraundorfer, and H.~Bischof.
\newblock Uav-based autonomous image acquisition with multi-view stereo quality
  assurance by confidence prediction.
\newblock {\em arXiv preprint arXiv:1605.01923}, 2016.

\bibitem{Mo:14}
C.~Mostegel, A.~Wendel, and H.~Bischof.
\newblock Active monocular localization: towards autonomous monocular
  exploration for multirotor mavs.
\newblock In {\em 2014 IEEE International Conference on Robotics and Automation
  (ICRA)}, pages 3848--3855, 2014.

\bibitem{NeJa:95}
T.~S. Newman and A.~K. Jain.
\newblock A survey of automated visual inspection.
\newblock {\em Computer Vision and Image Understanding}, 61(2):231--262, 1995.

\bibitem{Sa:15}
S.~A. Sadat, J.~Wawerla, and R.~Vaughan.
\newblock Fractal trajectories for online non-uniform aerial coverage.
\newblock In {\em 2015 IEEE International Conference on Robotics and Automation
  (ICRA)}, pages 2971--2976, 2015.

\bibitem{SchKo:12}
K.~Schmid, H.~Hirschm{\"u}ller, A.~D{\"o}mel, I.~Grixa, M.~Suppa, and
  G.~Hirzinger.
\newblock View planning for multi-view stereo 3d reconstruction using an
  autonomous multicopter.
\newblock {\em Journal of Intelligent \& Robotic Systems}, 65(1-4):309--323,
  2012.

\bibitem{ScRo:03}
W.~Scott, G.~Roth, and J.-F. Rivest.
\newblock View planning for automated 3d object reconstruction inspection.
\newblock {\em ACM Computing Surveys}, 35(1), 2003.

\bibitem{ScRoRi:03}
W.~Scott, G.~Roth, and J.-F. Rivest.
\newblock View planning for automated 3d object reconstruction inspection.
\newblock {\em ACM Computing Surveys}, 35(1), 2003.

\bibitem{Sc:09}
W.~R. Scott.
\newblock Model-based view planning.
\newblock {\em Machine Vision and Applications}, 20(1):47--69, 2009.

\bibitem{Sheinin_2016_CVPR}
M.~Sheinin and Y.~Y. Schechner.
\newblock The next best underwater view.
\newblock In {\em The IEEE Conference on Computer Vision and Pattern
  Recognition (CVPR)}, June 2016.

\bibitem{SuBa:98}
R.~S. Sutton and A.~G. Barto.
\newblock {\em Introduction to Reinforcement Learning}.
\newblock MIT Press, Cambridge, MA, USA, 1st edition, 1998.

\bibitem{TaATs:95}
K.~A. Tarabanis, P.~K. Allen, and R.~Y. Tsai.
\newblock A survey of sensor planning in computer vision.
\newblock {\em IEEE transactions on Robotics and Automation}, 11(1):86--104,
  1995.

\bibitem{Ta:95}
G.~H. Tarbox and S.~N. Gottschlich.
\newblock Planning for complete sensor coverage in inspection.
\newblock {\em Computer Vision and Image Understanding}, 61(1):84--111, 1995.

\bibitem{5597268}
M.~Trummer, C.~Munkelt, and J.~Denzler.
\newblock Online next-best-view planning for accuracy optimization using an
  extended e-criterion.
\newblock In {\em Pattern Recognition (ICPR), 2010 20th International
  Conference on}, pages 1642--1645, 2010.

\bibitem{4270361}
S.~Wenhardt, B.~Deutsch, E.~Angelopoulou, and H.~Niemann.
\newblock Active visual object reconstruction using d-, e-, and t-optimal next
  best views.
\newblock In {\em The IEEE Conference on Computer Vision and Pattern
  Recognition (CVPR)}, June 2007.

\bibitem{Wh:97}
P.~Whaite and F.~P. Ferrie.
\newblock Autonomous exploration: Driven by uncertainty.
\newblock {\em IEEE Transactions on Pattern Analysis and Machine Intelligence},
  19(3):193--205, 1997.

\end{thebibliography}
}

\end{document}